\documentclass{article}
\usepackage{spconf,amsmath,graphicx}
\usepackage{makecell}
\usepackage{diagbox}
\usepackage{tabularx}

\newcolumntype{C}{>{\centering\arraybackslash}X}

\begin{document}


\title{A-PixelHop: A Green, Robust and Explainable Fake-Image Detector}

\name{Yao Zhu\textsuperscript{1}, Xinyu Wang\textsuperscript{1},
Hong-Shuo Chen\textsuperscript{1}, Ronald Salloum\textsuperscript{2},
C.-C. Jay Kuo\textsuperscript{1}}

\address{Ming Hsieh Department of Electrical Engineering, University of Southern California$^1$\\
School of Computer Science and Engineering, California State University, San Bernardino$^2$}

\maketitle

\begin{abstract}

A novel method for detecting CNN-generated images, called Attentive PixelHop (or
A-PixelHop), is proposed in this work. It has three advantages: 1) low
computational complexity and a small model size, 2) high detection
performance against a wide range of generative models, and 3)
mathematical transparency.  A-PixelHop is designed under the assumption
that it is difficult to synthesize high-quality, high-frequency
components in local regions.  It contains four building modules: 1)
selecting edge/texture blocks that contain significant high-frequency
components, 2) applying multiple filter banks to them to obtain rich
sets of spatial-spectral responses as features, 3) feeding features to
multiple binary classifiers to obtain a set of soft decisions, 4)
developing an effective ensemble scheme to fuse the soft decisions into the
final decision. Experimental results show that A-PixelHop outperforms
state-of-the-art methods in detecting CycleGAN-generated images.
Furthermore, it can generalize well to unseen generative models and datasets. 

\end{abstract}
\begin{keywords}
image forensics, fake-image detection, neural networks, generative models
\end{keywords}
\section{Introduction}\label{sec:intro}

In recent years, there has been a rapid development of image-synthesis
techniques based on convolutional neural networks (CNNs), such as generative adversarial networks
(GANs) \cite{goodfellow2014generative}.  Such techniques have demonstrated the ability to generate high-quality fake images, and as a result, have raised concerns that it will become increasingly challenging to distinguish fake (or synthetic) and real
(or authentic) images. Determining whether an image was synthesized by a
specific CNN-based architecture is relatively straightforward. This can be accomplished by training a
classifier using real and fake images generated by the specific
CNN-based architecture. However, there exist many different fake-image generators, and thus, it is essential to develop a generic detection
method that can generalize well to fake images generated by a wide range
of generative models. This is the objective of our current research.

Most state-of-the-art methods for detection of CNN-generated images are
based on deep neural networks. Different architectures have proven to be effective in detecting fake images. However, deep-learning-based detection
methods need an enormous amount of data to maintain good performance. Because of the rapid evolution of image-synthesis techniques, training datasets from multiple generative models and/or extensive data augmentation are needed in order for these detection methods to generalize well to unseen generative models.

In contrast to deep-learning-based methods, a novel detector based on signal
processing, called Attentive PixelHop (or A-PixelHop), is proposed in
this work. It has three key characteristics: low computational and memory
complexity (i.e., green), high detection performance against a wide range
of generative models (i.e., robustness), and mathematical transparency
(i.e., explainability). Its design is based on the assumption that
high-quality, high-frequency components in local regions are more
difficult to generate. 

A-PixelHop has four building modules. Its first module selects
edge/texture blocks that contain significant high-frequency components.
Its second module applies multiple filter banks to them to obtain rich
sets of spatial-spectral responses as features. Its third module feeds
features to multiple binary classifiers to obtain a set of soft
decisions. Its last module adopts an effective ensemble scheme to fuse
the soft decisions into the final decision. It is demonstrated by
experimental results that A-PixelHop outperforms state-of-the-art methods for CycleGAN-generated images. Furthermore, it is demonstrated that
A-PixelHop can generalize well to unseen generative models and datasets. 

\begin{figure*}[htb]
\centering
\centerline{\includegraphics[width=\textwidth]{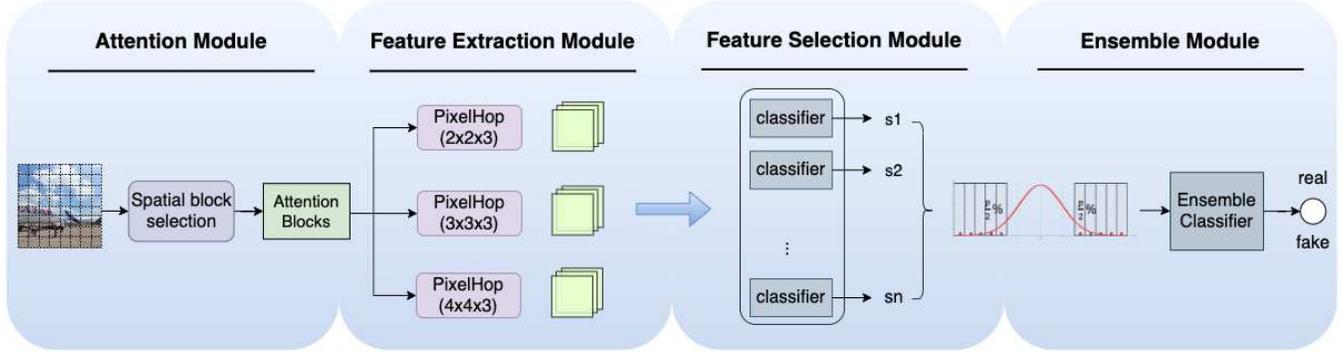}}
\vspace{-2mm}
\caption{The block-diagram of the A-PixelHop method.}\label{fig:system}
\end{figure*}

\section{Related Work}\label{sec:relatedwork}

\textbf{Detecting CNN-generated Images.} In this paragraph, we provide a brief summary of existing methods for detection of CNN-generated images. Inspired by solutions in steganalysis,
Cozzolino \cite{cozzolino2017recasting} proposed a CNN architecture that
mimics the rich models in feature extraction and classification.  Zhang \textit{et al.} \cite{zhang2019detecting} proposed to feed spectral input (rather than pixel input) to a classifier.  They also introduced a GAN
simulator, called AutoGAN, that simulates artifacts produced by popular GAN models.  Recently,
Wang \textit{et al.} \cite{wang2020cnn} trained a
classifier with a large number of ProGAN-generated images and evaluated it on images synthesized by eleven different generators. Their work
showed the effectiveness of extensive data augmentation in improving the generalization ability of a classifier. 

\textbf{Subspace Learning.} Our work follows the methodology of
subspace learning. Although subspace learning has a long history, Kuo
{\em et al.} \cite{kuo2019interpretable} built a link between subspace
learning and the convolutional operations in CNNs recently. A set of
convolutional filters in a given convolutional layer of a CNN can be
interpreted as a set of filters in one filter bank. The filter
parameters in CNNs are obtained via end-to-end optimization through
back-propagation. However, in subspace learning, filter parameters are derived by the
statistical analysis of pixel correlations inside a local region covered
by the filter. For example, for a filter of size 5x5x3, where 5x5 corresponds to the spatial window size and 3 corresponds to the three color channels, we would examine
the correlations between the 5*5*3 = 75 pixels. A variant of principal
component analysis (PCA), called the Saab (Subspace approximation via
adjusted bias) transform, was introduced in \cite{kuo2019interpretable}
and used to determine filter parameters. The concept of multiple
convolutional layers can be ported to subspace learning, leading to
successive subspace learning (SSL), and the corresponding architecture
is called the PixelHop \cite{chen2020pixelhop}. PixelHop offers an
unsupervised and feedforward feature learning process. Neither
back-propagation nor labels are needed in deriving filter parameters. SSL
has been applied to image classification \cite{chen2020pixelhop} and 3D
point cloud classification \cite{zhang2020pointhop}, among many others.
In image forensics, DefakeHop \cite{chen2021defakehop} was proposed to
detect deepfake videos. 

It is worthwhile to emphasize that our work is different from DefakeHop
in two main aspects. First, DefakeHop used facial landmarks to crop out
eyes, nose and mouth regions and perform detection in each region. However, in this work, we need to consider generic fake images and cannot rely on the special
facial regions here. Second, DefakeHop leveraged cascaded PixelHop
units; it belongs to the category of SSL. On the other hand, our work adopts multiple
single-stage PixelHop units in parallel, and thus, belongs to the category of
parallel subspace learning (PSL).

\section{Proposed A-PixelHop Method}\label{sec:proposedmethod}

\begin{figure}[htb]
\centering
\centerline{\includegraphics[width=8.5cm]{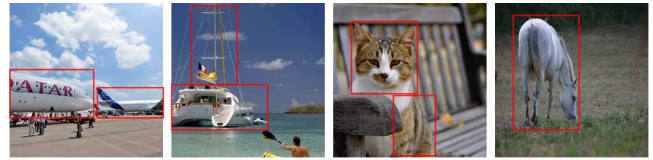}}
\caption{Illustration of selected spatial regions that contain complex
and fine details.} \label{fig:attention}
\vspace{-3mm}
\end{figure}

An overview of the proposed A-PixelHop method is given in Fig.
\ref{fig:system}.  It takes authentic or CNN-generated fake images
as input and generates a binary decision - true or fake.  Its four
modules are elaborated below. 

\noindent
{\bf 1) Spatial-Block Selection.} An image is first partitioned into
non-overlapping blocks of size 16x16. Under the assumption that it is
more difficult for CNN-based generators to synthesize high-frequency
components in images, the spatial attention module is developed to
select blocks that contain complex and/or fine details. There are many
ways to implement this idea. Here, we remove the DC component of the
block, compute variances of block residuals,
and select blocks that have a larger partial sum of variances. These
blocks correspond to edge/textured regions in images. Examples of
selected spatial regions are
shown in Fig. \ref{fig:attention}

\noindent
{\bf 2) Parallel PixelHop.} A PixelHop unit consists of a set of filters
of the same size that operate on all pixels in a block in parallel. A
filter is a 3D tensor of size $(s,s,c)$, where $s$x$s$ are the spatial
dimensions and $c$ is the spectral dimension.  Typically, $s=2,3,4$ and
$c=3$ for color images. We employ multiple PixelHop units in parallel for feature
extraction to increase feature diversity. Filter weights are determined
by a variant of PCA called the Saab transform 
\cite{kuo2019interpretable}.  As shown in Fig. \ref{fig:system}, we use
three parallel PixelHop units for feature extraction, where the filters are
of sizes 2x2x3, 3x3x3 and 4x4x3, respectively. For filters of sizes
$(s,s,c)$, there are $s^2 c$ channels and each channel has $(17-s)^2$
spatial responses. 

\begin{figure}[htb]
\begin{minipage}[b]{.45\linewidth}
  \centering
  \centerline{\includegraphics[width=4.2cm]{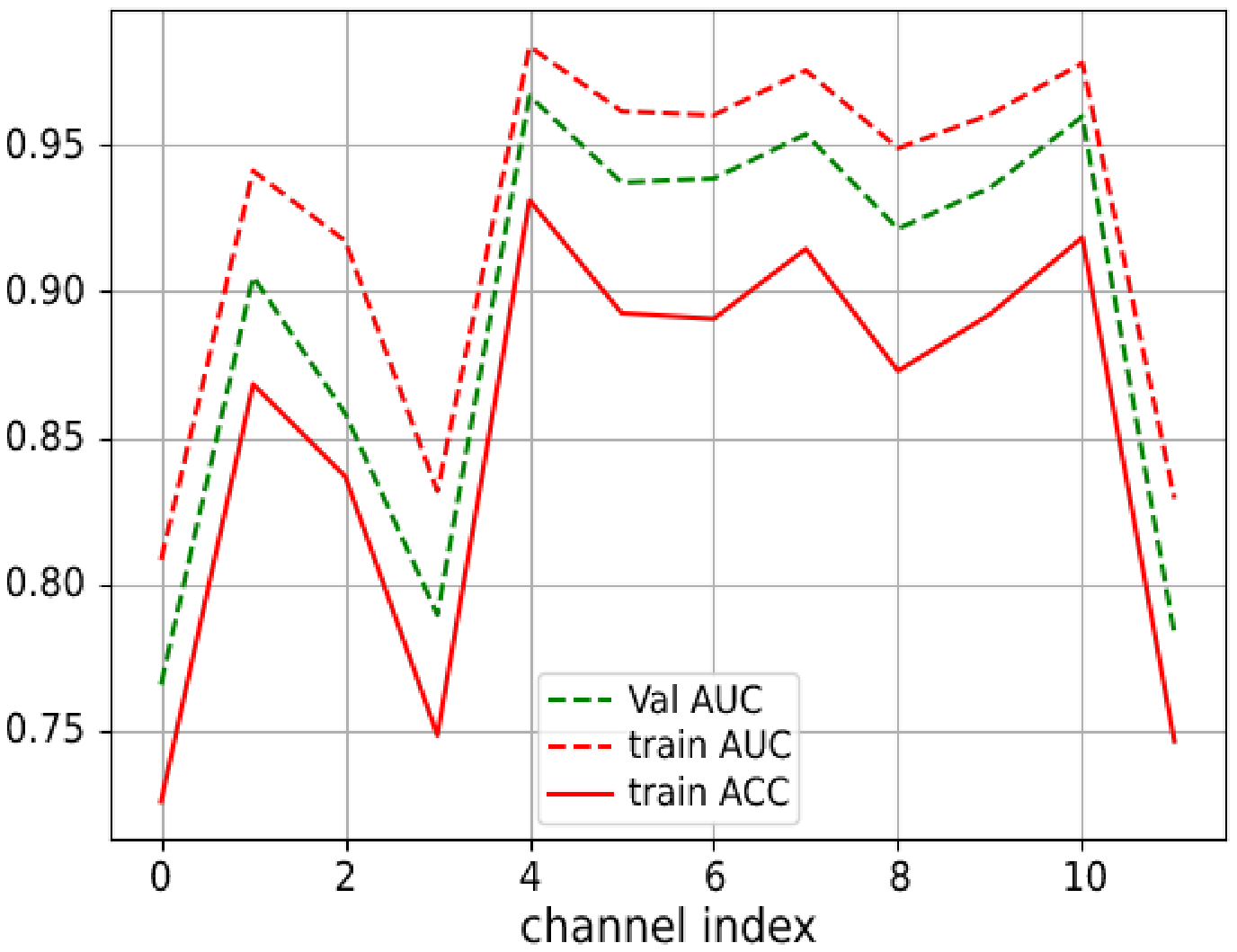}}
  \centerline{ap2or/2x2x3}\medskip
\end{minipage}
\hfill
\begin{minipage}[b]{.50\linewidth}
  \centering
  \centerline{\includegraphics[width=4.2cm]{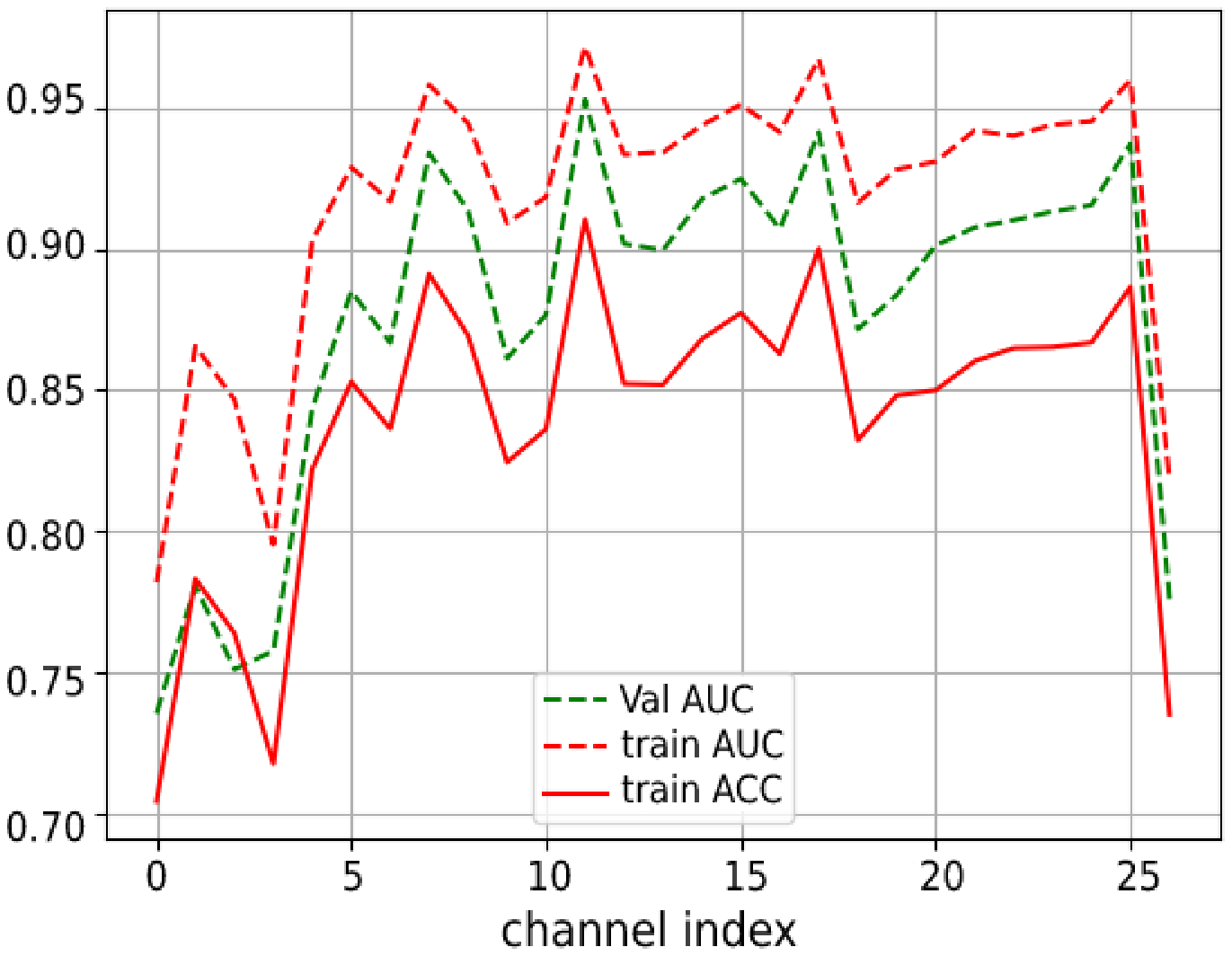}}
  \centerline{ap2or/3x3x3}\medskip
\end{minipage}

\begin{minipage}[b]{.45\linewidth}
  \centering
  \centerline{\includegraphics[width=4.2cm]{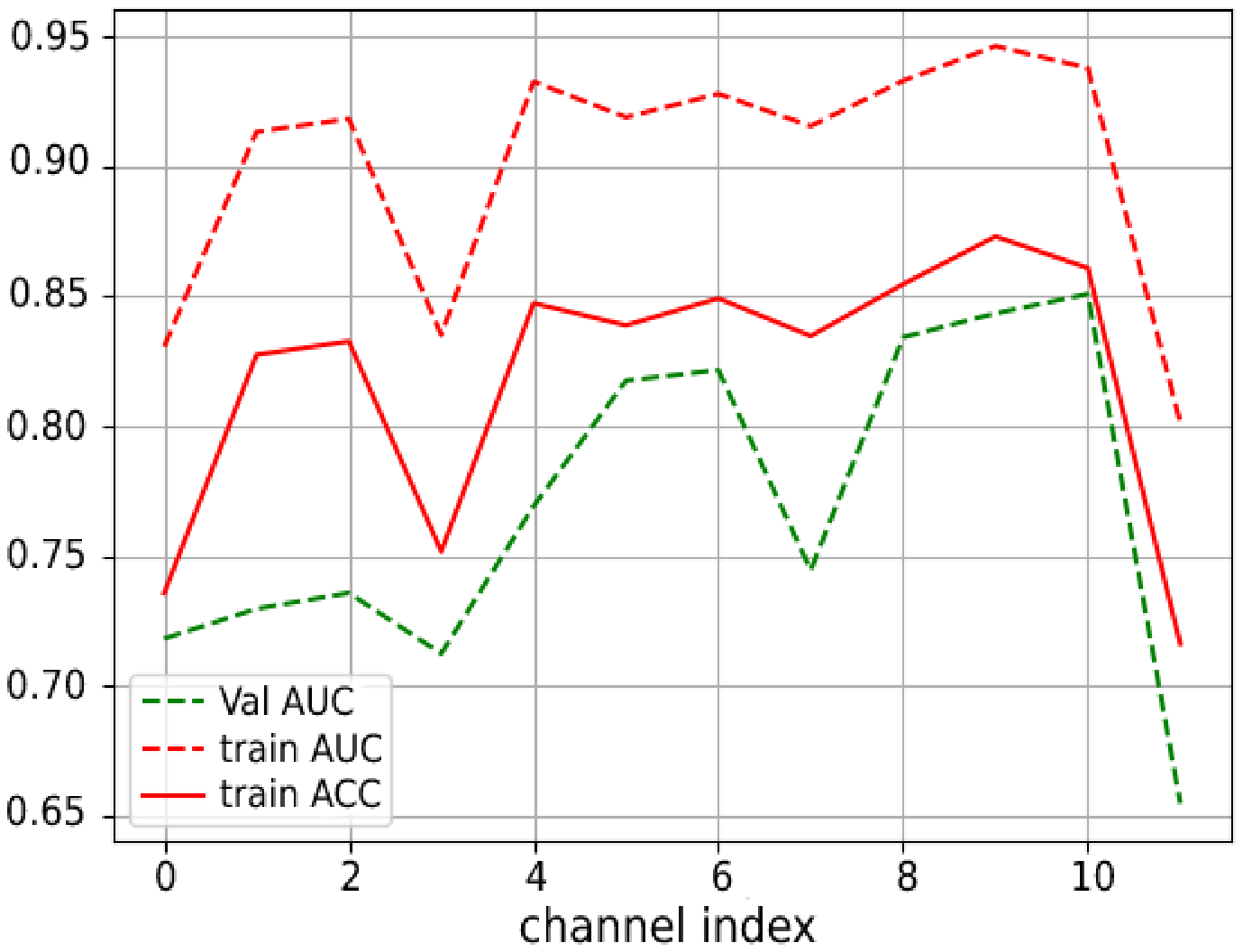}}
  \centerline{ProGAN/2x2x3}\medskip
\end{minipage}
\hfill
\begin{minipage}[b]{.48\linewidth}
  \centering
  \centerline{\includegraphics[width=4.2cm]{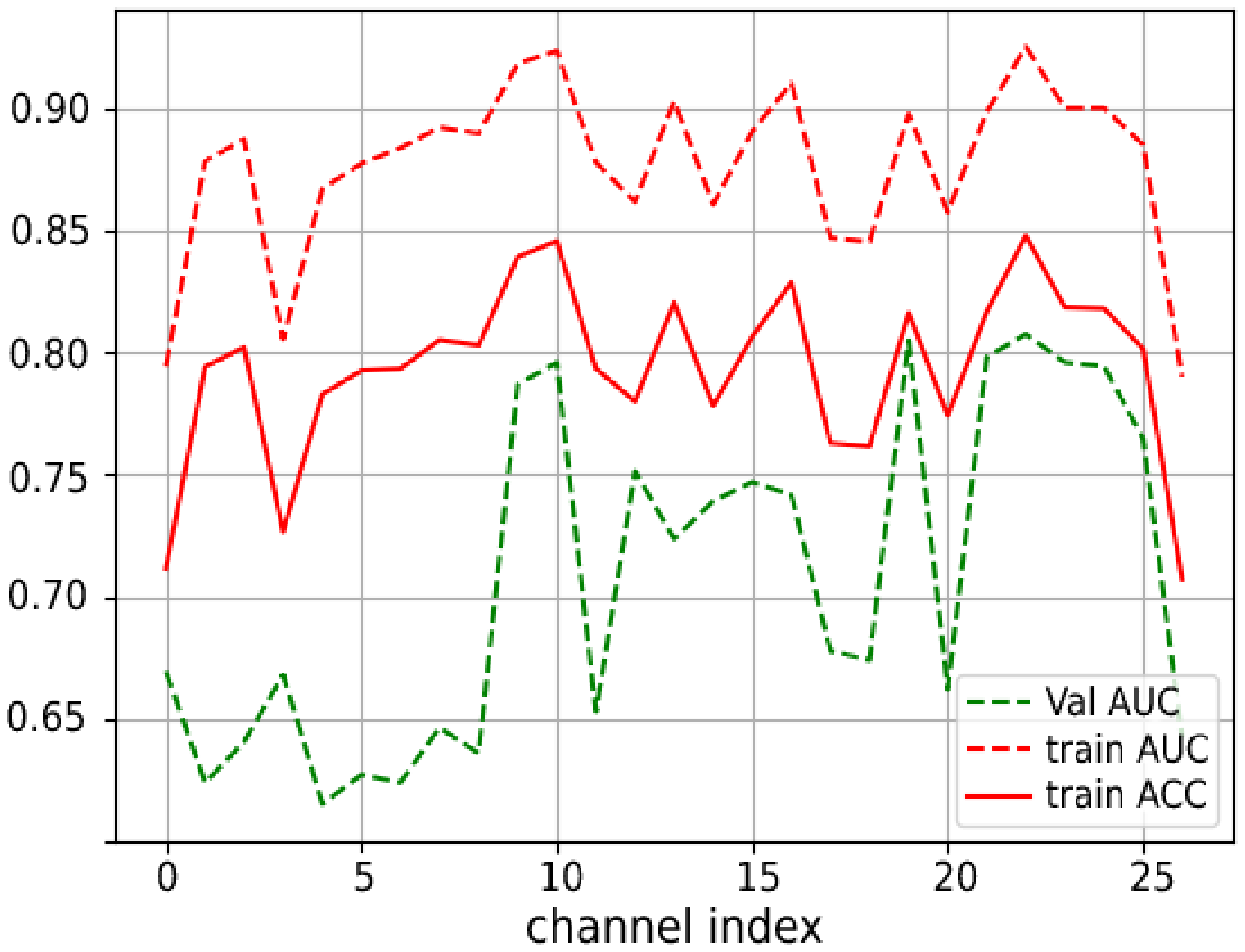}}
  \centerline{ProGAN/3x3x3}\medskip
\end{minipage}
\vspace{-3mm}
\caption{AUC/ACC performance of channel-wise classification using
filters of size 2x2x3 and 3x3x3, for the apple-orange subset under CycleGAN
(top two rows) and ProGAN (bottom two rows).} \label{fig:cwres}
\end{figure}

\noindent
{\bf 3) Classification and Discriminant Channel Selection.} We use
channel-wise spatial responses to train an XGBoost classifier to select
discriminant channnels.  The channel-wise classification performance
curves measured by the area-under-the-curve (AUC) and the accuracy
(ACC) for the apple vs. orange subset in the CycleGAN \cite{isola2017image} and 
ProGAN \cite{karras2017progressive} datasets are shown in Fig. \ref{fig:cwres}. We observe a consistent
trend between the validation AUC and training AUC curves and can select a
couple of discriminant channels based on the peaks of the validation AUC
curves. In the experiments, we select two and three optimal filters from
each PixelHop unit for CycleGAN and ProGAN, respectively. As a result,
we have six and nine discriminant filters for CycleGAN and ProGAN, respectively, and use their soft decision
scores from the XGBoost classifier as features for the image-level
decision ensemble. 

\begin{figure}[htb]
  \centering
  \centerline{\includegraphics[width=7cm]{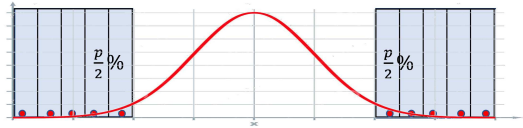}}
  \vspace{-3mm}
\caption{Image-level ensemble based on soft decisions selected from two ends of 
the distribution.}\label{fig:twoend}
\end{figure}

\vspace{-3mm}
\noindent
{\bf 4) Image-Level Ensemble.} A-PixelHop makes the final
image-level decision by ensembling block-level soft decisions.  However,
not all soft decisions are equally important. Soft decisions close to
the center (namely, 50\% vs. 50\%) are not as discriminant as those
lying at two ends. By following this line of thought, we sample $0.5 p
\%$ soft decisions at two ends of the distribution as shown in Fig.
\ref{fig:twoend}, where representative soft decisions (denoted by red
dots) are selected. As a result, only $p\%$ of representative soft
decisions are fed to the image-level ensemble classifier as features.
Note that a typical $p$ value is 10, 20 or 30.  For images of
size 256x256, it is fine to simply select the top and bottom $0.5 p
\%$ soft decisions without sampling. However, for images of
higher resolution (e.g. 3000x4000), the number of discriminant blocks
is very large if $10 \leq p \leq 30$. On the other hand, if we set $p$ to a
small value (say, $p=1$), the selected samples are likely to be outliers
and they are not representative enough. Thus, a sampling scheme offers a
good balance between representation and discrimination. 

\vspace{-7mm}
\begin{table*}[!h]
\begin{tabularx}{\textwidth}{c|CCCCCCCCCC|c} \hline
Methods & ap2or & ho2zeb & win2sum & citysc. & facades & map2sat & Ukiyoe & VanGogh & Cezanne & Monet & ave.\\ 
\hline \hline
DenseNet & 79.1 & 95.8 & 67.7 & 93.8 & 99.0 & 78.3 & 99.5 & 97.7 & 99.9 & 89.8 & 89.2\\
XceptionNet & 95.9 & 99.2 & 76.7 & 100.0 & 98.6 & 76.8 & 100.0 & 99.9 & 100.0 & 95.1 & 94.5\\
Cozzolino2017 & 99.9 & 99.9 & 61.2 & 99.9 & 97.3 & 99.6 & 100.0 & 99.9 & 100.0 & 99.2 & 95.1 \\
Auto-Spec & 98.3 & 98.4 & 93.3 & 100.0 & 100.0 & 78.6 & 99.9 & 97.5 & 99.2 & 99.7 & 97.2 \\ 
\hline
Ours & 99.2 & {99.8} & {100.0} & {94.4} & {100.0} & 94.1 & {100.0} & {100.0} & {100.0} & 99.4 & \textbf{98.7}\\
\hline
\end{tabularx}
\vspace{-3mm}
\caption{Comparison of test accuracy of fake-image detectors against 10 CycleGAN subsets in 
Experiment I. Performance numbers for DenseNet, XceptionNet, and Cozzolino2017 are taken from \cite{marra2018detection}. \textbf{Boldface} is used to indicate best performance.}\label{table1:loo}
\end{table*}

\begin{table*}[!h]
{\begin{tabularx}{\textwidth}{c|CCCCCCCCCCC | C }
\hline
Methods & Pro-GAN & Style-GAN & Big-GAN & Cycle-GAN & Star-GAN & Gau-GAN & CRN & IMLE & SITD 
& SAN & Deep-fake & mAP\\ \hline\hline
Auto-Spec & 75.6 & 68.6 & 84.9 & 100.0 & 100.0 & 61.0 & 80.8 & 75.3 & 89.9 & 66.1 & 39.0 & 76.5\\ 
Wang {\em et al.} & 100. & 96.3 & 72.2 & 84.0 & 100. & 67.0 & 93.5 & 90.3 & 96.2 & 93.6 & 98.2 & 90.1 \\\hline
Ours ($10\%$) & 99.9 & {99.9} & 77.1 & {97.8} & 100.0 & {94.6} & 76.4 & 92.8 & 76.5 & 84.4 & 94.5 & 90.4\\
Ours ($20\%$) & 99.9 & {99.9} & 75.2 & {97.3} & 100.0 & {94.7} & 86.8 & 95.3 & 76.5 & 83.8 & 93.3 & 91.2\\
Ours ($30\%$) & 99.9 & {99.9} & 74.4 & {96.7} & 99.9 & {94.8} & 91.3 & 98.2 & 76.7 & 80.3 & 91.3 & \textbf{91.2}\\ \hline
\end{tabularx}}{}
\vspace{-3mm}
\caption{Comparison of the average precision of fake-image detectors against eleven generative 
models in Experiment II. \textbf{Boldface} is used to indicate best performance.}\label{table2:cross}
\vspace{-3mm}
\end{table*}

\section{Experiments}\label{sec:typestyle}
\vspace{-3mm}

In this section, we compare the performance of our proposed method with several existing methods in two different experimental setups. 

{\bf Experiment I.} In the first experiment, we utilize a dataset consisting of 10 subsets, where each subset
contains both real and CycleGAN-generated images
\cite{zhang2019detecting, isola2017image,marra2018detection}. For example, the
\textit{horse2zebra} subset includes real horse and zebra images for
training CycleGAN and corresponding fake horse and zebra images
generated from the trained model. It has 14 semantic categories,
including Apple, Orange, Horse, Zebra, Yosemite summer, Yosemite winter,
Facades, CityScape\_Photo, Satellite Image, Ukiyoe, Van Gogh, Cezanne,
Monet and Photo. There are over 36K images in this dataset. 

We compared the proposed A-PixelHop method with several state-of-the-art methods, including Cozzolino2017 \cite{cozzolino2017recasting} and AutoGAN
with spectral input (Auto-Spec) \cite{zhang2019detecting}. Here, we
follow the leave-one-out setting described in \cite{zhang2019detecting, marra2018detection}, where one subset is set aside for
testing while the other nine subsets are used in the training process. Table \ref{table1:loo} shows the test accuracy of the proposed method as well as the existing methods. We see that A-PixelHop reaches 100\% accuracy for
five (out of ten) subsets. Its average accuracy over the 10 subsets is
98.7\%, which is best among all methods. 

{\bf Experiment II.} This second experimental setup was utilized in \cite{wang2020cnn}
in order to evaluate how well a given detection method generalizes to unseen
generative models. The dataset used here contains images synthesized by a wide variety of generative
models. All of them have an
upsampling-convolutional structure.  In the training set, fake images from
20 object categories are generated by the ProGAN model only. There are
720K real/fake image pairs in the training set and 4K images in the
validation set. In the testing set, fake images are generated by the following eleven models: ProGAN \cite{karras2017progressive}, StyleGAN \cite{karras2019style}, BigGAN \cite{brock2018large}, CycleGAN \cite{isola2017image}, StarGAN \cite{choi2018stargan}, GauGAN \cite{park2019semantic}, CRN \cite{chen2017photographic}, IMLE
 \cite{li2019diverse}, SITD \cite{chen2018learning}, SAN
 \cite{dai2019second}, and Deepfake \cite{rossler2019faceforensics++}.


In this experiment, we follow the procedure specified in
\cite{wang2020cnn}, by first training A-PixelHop with real and
ProGAN-generated fake images, and then evaluating its detection performance on real images or fake images generated by the aforementioned eleven generative models.  The
performance comparison between A-PixelHop and two existing
methods, Auto-Spec and the method proposed by Wang {\em et al.} in \cite{wang2020cnn}, is shown in
Table \ref{table2:cross}. It is worthwhile to emphasize three points.
First, since we do not include augmentation in the training of
A-PixelHop, we compare against \cite{wang2020cnn}
under the no augmentation setting. Second, we evaluate the performance in terms of
average precision (AP) so as to be
consistent with \cite{wang2020cnn}. Third, we collect a total of
10\%, 20\% and 30\% of samples from two ends for the image-level ensemble
and show the corresponding mean AP (mAP) values. We see from the table that
A-PixelHop outperforms both Auto-Spec and the method from \cite{wang2020cnn} in
all three cases (i.e., 10\%, 20\% and 30\% of samples) in terms of mAP. A-PixelHop outperforms both
Auto-Spec and the method from \cite{wang2020cnn} by a large margin in the case of
Gau-GAN.  A-PixelHop performs worse in the case of Big-GAN, SITD and
SAN, indicating a weaker transferability from ProGAN to these three
generative models. It demands further exploration. SITD and SAN generate images of very high resolution (e.g., 3Kx4K pixels), which does not match well with that of the training images generated by ProGAN.
One possible fix is to rescale these large images to smaller ones in the
pre-processing step. 

Based on Experiments I and II, we conclude that the proposed A-PixelHop
method is robust in the sense that it generalizes well to different
semantic categories as well as unseen generative models. In addition, it offers state-of-the-art detection
performance. 

{\bf Model Size Computation.} We compare the model size (in terms of number of parameters) of the proposed
A-PixelHop method with that of the other methods in Table \ref{tab:model_size}.  Auto-Spec
and the method from \cite{wang2020cnn} utilized Resnet34 and Resnet50, respectively.
The method from \cite{cozzolino2017recasting} used a light weight CNN that has two
convolutional layers and one fully connected layer. The model has 1K
parameters. A-PixelHop has different model sizes in Experiments I and
II.  As shown in Table \ref{tab:apixelhop_size}, it selects 6 and 9
discriminant channels in Experiments I and
II, respectively, and trains one XGBoost classifier
for each channel. Each XGBoost classifer has 100 trees with a maximum
depth of 6 and has a model size of 19K. Furthermore, it
trains an ensemble XGBoost classifier that has 10 trees with depth equal
to one. Its model size is 40. 

\vspace{-2mm}
\begin{table}[!h]
\centering
    \begin{tabularx}{\linewidth}{c c c c c} \hline
        Exp. & Ours & Auto-Spec & Cozzolino2017 
        & Wang {\em et al.} \\ \hline
        I  & 114K & 21.8M & 1K & --  \\         
        II & 171K & 21.8M & -- & 25.6M\\        \hline
    \end{tabularx}
\vspace{-4mm}
\caption{Model size comparison (in terms of number of parameters).}\label{tab:model_size}
\end{table}
\vspace{-4mm}

\begin{table}[!h]
\centering
    \begin{tabularx}{\linewidth}{c| C ||c | C} \hline
    components      & para \# & components      & para \# \\ \hline
    2 (2x2x3)  & 24      & 3 (2x2x3) & 36 \\      
    2 (3x3x3)  & 54      & 3 (3x3x3) & 81 \\
    2 (4x4x3)  & 96      & 3 (4x4x3) & 144 \\    
    6 XGBoost & 6x19K    & 9 XGBoost & 9x19K \\
    1 XGBoost  & 40      & 1 XGBoost & 40 \\ \hline
    Total & 114K         &Total & 171K \\ \hline
    \end{tabularx}
\vspace{-4mm}
\caption{Model size computation for A-PixelHop for Experiment I 
(left) and II (right).}\label{tab:apixelhop_size}
\end{table}

\vspace{-5mm}
\section{Conclusion and Future Work}\label{sec:majhead}
\vspace{-2mm}

A green, robust, high-performance and explainable method, called
A-PixelHop, to detect CNN-generated fake images was presented in this
work. A-PixelHop used the filter-bank signal processing tool to extract
discriminant joint spatial-spectral components as features and fed them
to the XGBoost classifer to derive the block-level decision. Finally, it
adopted an ensemble learning tool to fuse multiple block-level soft
decisions to obtain the final image-level decision. The superior
performance of A-PixelHop was demonstrated by experimental results.  As
future extension, we plan to apply A-PixelHop to distinguish real/fake
images that are manipulated by other operations such as
compression, blurring, additive noise, etc. 

\vfill\pagebreak

\bibliographystyle{IEEEbib}
\bibliography{refs}

\end{document}